\documentclass[conference]{IEEEtran}
\IEEEoverridecommandlockouts
\usepackage{cite}
\usepackage{amsmath,amssymb,amsfonts}
\usepackage{algorithmic}
\usepackage{graphicx}
\usepackage{textcomp}
\usepackage{xcolor}
\usepackage[utf8]{inputenc}
\def\BibTeX{{\rm B\kern-.05em{\sc i\kern-.025em b}\kern-.08em
    T\kern-.1667em\lower.7ex\hbox{E}\kern-.125emX}}
\begin{document}

\title{Inferring the ground truth through crowdsourcing}

\author{\IEEEauthorblockN{Jean Pierre Char}
\IEEEauthorblockA{\textit{Student of M.Sc. Computer Science} \\
\textit{University Of Passau}\\
Passau, Germany \\
char01@gw.uni-passau.de}
}

\maketitle

\begin{abstract}
Universally valid ground truth is almost impossible to obtain or would come at a very high cost. For supervised learning without universally valid ground truth, a recommended approach is applying crowdsourcing: Gathering a large data set annotated by multiple individuals of varying possibly expertise levels and inferring the ground truth data to be used as labels to train the classifier. Nevertheless, due to the sensitivity of the problem at hand (e.g. mitosis detection in breast cancer histology images), the obtained data needs verification and proper assessment before being used for classifier training. Even in the context of organic computing systems, an indisputable ground truth might not always exist. Therefore, it should be inferred through the aggregation and verification of the local knowledge of each autonomous agent.\\
\end{abstract}


\section{Introduction}
In a supervised learning problem, it is imperative to have accurate labelling of the data to properly train a classifier. The indisputable ground truth describing this data is constructed by means of these labels. However, as will be illustrated in this article, in many scenarios, labelled data is not always available due to a variety of reasons. Also, even if it is accessible, it might not be \textit{"universally valid"} which makes it unreliable for training the classifier. This leads researchers to consider crowdsourcing as a method to label their data by relying on the \textit{"crowd"}\cite{Crowdsourcing}. Nevertheless, crowdsourcing is not as trivial as many would expect\cite{CharacterizationOfCrowdsourcingPractices}. Therefore, user verification and data assessment mechanisms need to be put into action in order to obtain such ground truth. 

In this paper, the concept of crowdsourcing is introduced in section 3, along with some services and application areas to demonstrate the widespread use of this approach. In section 4, the study case of the possibility of inferring the ground truth through crowdsourcing is presented and many examples are given as reasons for considering this approach. This is done by listing the advantages and disadvantages of crowdsourcing and taking into consideration how to design such tasks and assess their results. In section 5, an application scenario in organic computing systems is considered to demonstrate how crowdsourcing can be used to infer ground truth data. Finally, a conclusion is formulated and further suggestions are proposed in section 6.\\

\section{Related Work}
Over the last decades, a lot of research has been invested in supervised learning, its algorithms and their accuracy and performance depending on different data sets\cite{ComparisonOfSupervisedLearningAlgorithms}.
Nevertheless, many obstacles have been encountered, such as the abundance of cheap unlabelled data, the high cost and difficulty of labelling it and the subjectivity of individuals in languages\cite{WordSenseAndSubjectivity} among many others.

In the active learning scenario\cite{ActiveLearning}, researchers usually rely on a user (often called oracle), who knows the ground truth, to label the abundantly existing unlabelled data. This oracle is an expert in the field and is hopefully reachable at a not so expensive cost. However, finding such an oracle might prove to be infeasible, i.e. he/she does not exist (or might be very expensive), which led to the rise of crowsdourcing\cite{Crowdsourcing} (also called active learning from crowds\cite{ActiveLearningFromCrowds}). This method allows them to transform unlabelled data into a set of data points where each one is labelled multiple times by multiple annotators.

Nevertheless, labels collected through crowdsourcing cannot be trusted blindly. This is due to people gaming crowdsourcing systems\cite{AMTScreening} and due to the fact that this kind of systems depends on the knowledge of the participants. Thus, in many cases, insufficient knowledge can lead to wrong answers being given involuntarily depending on the nature and topic of the task\cite{CharacterizationOfCrowdsourcingPractices}.\\

\section{Crowdsourcing}
\subsection{Term definition}
Crowdsourcing is a type of collective activity in which a particular individual, an organization (possibly non-profit) or a company proposes to a \textit{crowd} of individuals of varying knowledge, heterogeneity, and number, via flexible open call, the voluntary undertaking of a task\cite{Crowdsourcing}. However, accomplishing such tasks can also be rewarded with monetary compensation. The term ”Microwork” is used to describe paid work to accomplish these time-consuming tasks.\\

\subsection{Crowdsourcing services}
Crowdsourcing can be achieved through a variety of data collection methods, such as:
\begin{itemize}
    \item Compulsory web security tasks, such as reCAPTCHA\cite{recaptcha}.
    \item Free games, such as Listen Game\cite{ListenGame} or Peekaboom\cite{Peekaboom}.
    \item Paid work on online platforms, such as Amazon Mechanical Turk (AMT)\footnote{Amazon Mechanical Turk, https://www.mturk.com, last accessed on July 31, 2018.} or Figure Eight\footnote{Figure Eight (previously known as CrowdFlower, rebranded in 2018), https://www.figure-eight.com/, last accessed on July 31, 2018.}.
    \item Public annotation tools, such as LabelMe\footnote{LabelMe, http://labelme.csail.mit.edu/, last accessed on July 31, 2018.}.
    \item Online user survey software packages, such as SurveyMonkey\footnote{SurveyMonkey, https://www.surveymonkey.com/, last accessed on July 31, 2018.}.
    \item Online marketplaces for creative services, such as crowdSPRING\footnote{crowdSPRING, https://www.crowdspring.com/, last accessed on July 31, 2018.}.
\end{itemize}
As a counterpart, CrowdTruth\footnote{CrowdTruth framework, http://www.crowdtruth.org, last accessed on July 31, 2018.} was implemented. It serves as a library to process crowdsourcing results from Amazon Mechanical Turk and Figure Eight following the CrowdTruth methodology\cite{CrowdTruthMethadology}.\\

\subsection{Crowdsourcing application areas}
Since the rise of crowdsourcing, there has been many instances where it proved to be the solution to the problem at hand, such as:
\begin{itemize}
    \item Facebook turned to its users to transform the content of its online social platform from English to different languages. A crowd of generous users worked together translating it bits by bits while others confirmed the completeness of their work\cite{FacebookToGerman2008}.
    \item Goldcorp of Canada had complications locating exact spots of gold on its lands and decided to turn to the public for help. It made its exploration databases available to the public to go through and offered a prize to any individual or group of individuals who could tell the company where to find gold\cite{MassCollaboration}.
    \item The free, online, community encyclopedia Wikipedia with openly editable content\footnote{Wikipedia page on Wikipedia, https://en.wikipedia.org/wiki/Wikipedia, last accessed on July 31, 2018.}.
    \item Zooniverse\footnote{Zooniverse, https://www.zooniverse.org/, last accessed on July 31, 2018.}, a "citizen science", i.e. people-powered research, online platform seeking to advance scientific research through the participation of volunteers\cite{Zooniverse}.
\end{itemize}

\begin{figure}
    \centering
    \includegraphics[scale=0.6]{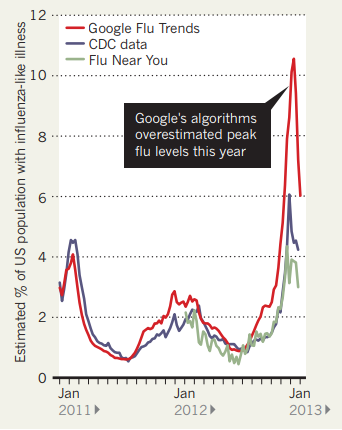}
    \caption{A comparison of three different methods of measuring the proportion of the US population with an influenza-like illness\cite{WhenGoogleGotFluWrong}}
    \label{fig:GoogleMessedUp}
\end{figure}

Nevertheless, crowdsourcing has failed in some instances also. For instance, when Google Flu Trends, a web-based method developed by Google for tracking seasonal flu using flu-related Internet searches, overestimated peak flu levels in the US for the end of 2012/start of 2013 as shown in Fig. \ref{fig:GoogleMessedUp} \cite{WhenGoogleGotFluWrong}. However, this was due to a bug in Google's algorithms that process user Internet searches to assess flu levels.\\

\section{Study Case: Possibility of inferring the ground truth through crowdsourcing}
\subsection{Reason: Problem of finding a universally valid ground truth}
As said before, in a supervised learning problem, it is expected to have labelled (i.e. ground truth) data to train the classifier. However, this is not always the case. In some cases, the data set might not be entirely labelled and in other cases, there might be many, possibly conflicting, labels classifying the same data point. This is mainly due to a complicated notion of truth where a single right annotation might not exist.

In many scenarios, a lack of universally valid ground truth can be noticed, such as:
\begin{itemize}
    \item In natural language processing applications where one word can have multiple meanings and not all people express their opinions, emotions and sentiments in text in the same manner\cite{WordSenseAndSubjectivity}.
    \item Emotion detection of characters in videos or images where subjectivity plays a crucial role. Even experts would provide contradicting opinions when judging what emotions are being experienced by the same character in the same scene\cite{InteractiveLearningWithoutGoundTruth}.
    \item Quantifying the beauty of outdoor places where natural features or even man-made structures may lead to places being classified as more scenic. It is an ethereal measure that is difficult to evaluate due to its subjective nature\cite{QuantifyingTheBeautyOfOutdoorPlaces}.
    \item Detecting small volcanoes in Magellan SAR images of Venus where experts may visually examine the images and provide a subjective noisy estimation of the truth\cite{SubjectiveLabellingOfVenusImages}.
    \item Mitosis detection in breast cancer histology images where expertise plays a big role and even experts might not agree on the same verdict\cite{IEEEBreastCancer}.
\end{itemize}
This lack of a universally valid ground truth or the difficulty of obtaining it has generated a large need for crowdsourcing.

\subsection{Crowdsourcing criteria}
Considering that crowdsourcing relies on anonymous individuals from the crowd, labels are not entirely reliable. As an attempt to reduce inaccurate labelling, crowdsourcing platforms and entities seeking crowdsourced labels impose constraints and acceptance criteria. Thus, an annotator profile might be reviewed by the crowdsourcing platform before being able to contribute in any way as done by Amazon Mechanical Turk. It might also be necessary for the participant to meet some additional criteria set by the entity seeking crowdsourced data. These conditions can be:
\begin{itemize}
    \item Age: Surveys targeting people in a specific age range, such as 18-24 years old individuals.
    \item Gender: Surveys targeting men or women for a specific topic, e.g. violence against women.
    \item Country of residence: Surveys targeting people living in a specific city, district, country or continent.
    \item Nationality: Surveys targeting people of a certain nation.
    \item Accuracy in previously completed tasks: Trustworthy and dedicated participants are obviously more desirable.
    \item Marital status: Surveys targeting single, married, divorced or widowed individuals.
    \item Employment status: Surveys targeting unemployed, self-employed or retired individuals, or part-time or full-time employees. 
    \item Profession: Surveys targeting university students, engineers, teachers, etc.
    \item Income level: Surveys targeting low income, middle income or high income individuals.
    \item Number of children: Surveys targeting individuals not having any children, having 1 child or multiple children.\\
\end{itemize}

\subsection{Advantages of crowdsourcing}
Compared to traditional recruitment methods, such as university participants pools, crowdsourcing platforms offer a large group of individuals willing to work for free (where the incentive is just entertainment) by playing games like Peekaboom\cite{Peekaboom} or for a median wage of \$1.38/hour on Amazon Mechanical Turk\cite{PaidCrowdsourcing}. This participants pool is usually very diverse (gender, age, education, income level, motivation and relevant experience), of different nationalities and fast to respond to newly offered tasks\cite{AMTExperiments, LimitationsOfCrowdsourcing}. Although, this population does not have direct ways for interacting on these platforms, worker discussion boards have been established that facilitate worker interaction, such as mturkforum.com or turkernation.com, and communities to share information can be found in totally independent online networks, such as Reddit or Facebook. Moreover, plug-ins have been developed that allow workers to carry out the tasks offered by their favored requesters, such as turkopticon.differenceengines.com or turkalert.com. As a result, some workers may know more about the tasks accessible to them and about the entities who posted them than is commonly expected. This can increase the portrayal of these workers in a sample and, perhaps, even contribute in workers having foresight of the study\cite{AMTWorkers}.\\

\subsection{Limitations of crowdsourcing}
It is crucial to understand the limitations of this form of data collection. As with any method involving monetary compensation, the motive of each participant may be questionable. It is possible that paying much less than the median reward would result in fewer participants taking part in the task, while paying much more would attract individuals who are not truly interested in completing the survey in good-faith. Alternatively, paying less may also convey to users a feeling that they are less bound to provide helpful responses, probably resulting in low-quality data\cite{LimitationsOfCrowdsourcing}. Some have also called this type of platforms failures for reasons, such as Amazon Mechanical Turk being skeptically spammy and allowing too many fraudulent users, requiring a lot of human intervention and effort for judging work as authentic or fake. In addition, the low rates, where most of them are around \$0.01/task, would not encourage participants to work in good-faith\cite{AMTFailure}.

On the other hand, voluntary tasks, if not advertised to the right group (group of people interested in the topic of the task)\cite{TargetingTheRightGroup}, might not receive enough participants. A low participation rate would only generate a skeptical set of low quality labels (possibly incomplete) which might harm the classifier training.\\

\subsection{Task design \& results assessement}
Labels gathered through crowdsourcing will most likely not be consistent due to:
\begin{itemize}
    \item Subjectivity of the participants which cannot be easily estimated or controlled\cite{WordSenseAndSubjectivity}.
    \item The high probability that some users might "game" the crowdsourcing system as shown in previous experiements\cite{AMTUserStudies, AMTScreening}.
    \item The lack of knowledge of some individuals leading them to involuntarily give wrong responses\cite{CharacterizationOfCrowdsourcingPractices}.
\end{itemize}
Therefore, methods have been established to assess participants and the correctness of their labels.

First of all, it is necessary to implement a reliable method of screening participants to remove the subset of those gaming the system\cite{AMTScreening}. There are some recommendations to consider when designing crowdsourcing tasks\cite{AMTUserStudies}, such as:
\begin{itemize}
    \item  Having one or multiple explicitly verifiable questions, i.e. gold questions whose responses are already known\cite{MajorityVoting}, as part of the task. This can be easily done in surveys by adding a question to select a specific answer out of multiple choices which would prove if the user is at least reading the task or not. Another important role of verifiable questions is in signaling to users that their responses will be validated, which may play a role in both the reduction of invalid responses and the increase of time spent on accomplishing the task\cite{AMTUserStudies}.
    \item Designing the task such that completing it accurately and in good-faith requires as much or less effort than random and fast completion\cite{AMTUserStudies}. 
    \item Having multiple ways to detect suspect responses. Even for highly subjective responses, there are certain patterns that, when put together, can signal a response as an attempt to game the system and make fast and easy money. For example, extremely short times spent on tasks and responses that are very repetitive across multiple tasks are flags of doubtful edits\cite{AMTUserStudies}.
    \item Understanding the classic measures of medical test theory: sensitivity and specificity. They are used to measure the performance of a binary classification test\cite{SensitivityAndSpecificity} and their importance in tasks that are formulated based on skewed data sets, such as "is there a dog in the picture?". Smart spammers can trick the system and take advantage of the task design and gain highly accurate results by just constantly giving the frequent class label, i.e. "there is no dog". In this case, their specificity would be high but their sensitivity would be nearly non-existing, since they hardly ever give the less frequent class label, i.e. "there is a dog"\cite{MajorityVoting}.
\end{itemize}

Second, the complexity of the task needs to be taken into consideration. Tasks are generally classified as simple, complex or creative\cite{CharacterizationOfCrowdsourcingPractices}, depending on many factors:
\begin{itemize}
    \item The topic of the task: A task, such as quantifying the beauty of outdoor places\cite{QuantifyingTheBeautyOfOutdoorPlaces}, is a creative task considering it depends on the opinion of the individual and his ability to express it. This kind of tasks can be very subjective, therefore increasing the complexity of assessing the results.
    \item Difficulty/simplicity of the words used to describe the task: Tasks that are described in an extremely scientific manner will not receive proper attention from non scientific users on crowdsourcing platforms.
    \item The level of knowledge required to answer the given task: A difficult task might require the participant to meet some educational qualifications, e.g. the user must have finished high school or is pursuing a university degree in a specific field.
    \item The number of possible choices in a task: Multiple possible choices would increase the complexity of a task and reduce the odds of obtaining the right answer in the end. For example, consider the complexity of a task where you have to evaluate the happiness of a character in a photo or a video. The possible choices might be either 1) "Sad" or "Happy" or 2) a scale from 1 to 10 where 1 is "Very Unhappy" and 10 is "Very Happy". 
    \item The size of the task, i.e. the time required to accomplish the task in good-faith: The longer the task is, the more the user has to invest in it which might eventually make him lose interest.
\end{itemize}
It's very beneficial to consider breaking down a complex task into multiple simple tasks in order to increase the number of different participants and reduce the possibility of losing the interest of the participant. For instance, a video annotation task (complex task) can be broken down into multiple image annotation tasks (simple tasks).

Third, in the case of microwork tasks, it is also essential to consider the dimensionality/reward ratio of the task. The longer and more difficult a task is, the more time and expertise it will require. A low-paid, long and demanding task will most likely not attract honest participants, if any at all. On the other hand, a high-paid, short and easy task will probably attract fraudulent users and therefore require more time to integrate mechanisms to detect deceitful responses to the given task.

Fourth, aggregating the obtained labels is not a trivial task. The threat of fraudulent participants and possibly wrong labels affecting the overall quality of the final labels still exists and can harm the classifier training. A naive approach like majority voting, a redundancy-based, task-oblivious approach where the aggregation of results from multiple workers for the same task is ultimately chosen as the final answer, e.g., by using averages or performing a majority vote, alone has its limitations, especially for higher ratios of spammers\cite{MajorityVoting}. Thus, other approaches might provide better results and need to be considered, such as:
\begin{itemize}
    \item Generative model of Labels, Abilities, and Difficulties (GLAD) in case the labellers are of unknown expertise to infer the expertise of each labeller, the difficulity of the task and an estimation of the actual label\cite{GLAD}.
    \item Weighted majority voting in case the labellers are of known varying expertise\cite{WeightedMajorityVoting}. This can be easily adopted to the example of mitosis detection in breast cancer histology images by taking doctors of varying expertise levels and correlating the weight of a doctor's label with his expertise level (e.g. a doctor with 20 years of experience would most likely be more accurate than a doctor with only 5 years of experience). The number of years of experience is just one parameter, many more can be used to estimate appropriate expertise levels of each doctor, such as certificates, number of clients, reputation, online reviews, etc. In the context of online crowdsourcing platforms, such as Amazon Mechanical Turk, expertise levels can be substituted for trust levels and number of previously completed tasks.
    \item Pruning low-quality labels, by removing low-quality labellers that tend to make errors, in order to increase label quality\cite{VoxPopuli}.
\end{itemize}

However, these approaches do not come without additional costs. The big advantage of the majority voting approach is its cheap cost and fast results. By designing a task with dependable methods of screening contributors and simplifying it to a reasonable extent, it is possible to reduce the ratio of dishonest and wrong answers drastically allowing the use of majority voting and obtaining high quality labels. Majority voting is the default approach in most survey results assessments where not much is known regarding each participant.

By following these design recommendations and choosing the proper assessment method depending on the task and the participants, researchers can obtain a \textit{close approximation} of the ground truth. This ground truth can then be utilized to train the classifier.\\

\section{Application scenario: Organic computing systems}

\subsection{Term definition}
Organic computing systems (or just organic systems) are systems whose design is inspired by nature. They profit from a combined presence of autonomy and self-organization. Therefore, they require low maintenance and employ a dynamic adaptation of the system behaviour to changing requirements of its operating environment. However, they also provide appropriate interfaces for potentially essential interaction with human agents or other entities on higher system levels, such as a goal change or urgent user intervention\cite{OrganicComputingBook}.\\

\subsection{Application scenario}
A lack of a universally valid ground truth can also be observed in the context of organic systems. In the case where researchers want to evaluate the entire state of the environment through the aggregation of the local knowledge of every autonomous agent of the system, the labels offered by the agents regarding the same section of the environment may not be consistent. This is due to many reasons, such as undetected faulty sensors, noise in the readings of the sensors, a software bug or even deceitful labels inserted by hackers.

In such a scenario, the expertise of each agent can be calculated using the following factors:
\begin{itemize}
    \item Last date of maintenance: A recently checked agent is less likely to produce errors than an agent whose last maintenance occurred a long time ago.
    \item Last date of encountered error: An agent that encountered an error recently is more likely to encounter new errors than an error-free agent, especially if the error was not properly addressed.
    \item Date of deployment into the environment: All hardware components have specific pre-defined lifetimes. Therefore, older sensors are more likely to be faulty than newly installed ones.
    \item Rating/quality of the equipment (e.g. sensor) of the agent: The higher the quality of an equipment, the better its performance.
\end{itemize}

Once the expertise of each agent is calculated, a weighted majority voting can be used to properly infer the ground truth from the conflicting labels of the agents. However, this is all under the assumption that there was no "gaming" of the system by hacking one or multiple agents. If that is the case, gold questions can be used, such as asking for an encrypted personal identification number (PIN) that the hacker would not easily know.\\

\section{Conclusion \& further suggestions}
A definition and multiple aspects of crowdsourcing have been presented. Design recommendations of crowdsourcing tasks and multiple methods to assess their results have been proposed in order to obtain a close approximation to the ground truth. Furthermore, an application scenario in the context of organic computing systems has been given to demonstrate the use of a design recommendation and a method to assess the crowdsourcing results. 

However, crowdsourcing can be expensive (to properly design tasks, assess their results and, if necessary, pay for participants). Alternatively, it would be worth considering the use of unsupervised learning. In that approach, there is no need for any kind of ground truth which would solve the problem at hand. Still, this might come at the cost of more time needed to detect the hidden structure of the data, accuracy and/or performance of the classifier. And as a compromise, semi-supervised learning may be used by using crowdsourcing to label a small portion of the data set and keeping the remainder of the data set unlabelled. \\

\end{document}